# Variational Chernoff Bounds for Graphical Models

**Pradeep Ravikumar**
School of Computer Science
Carnegie Mellon University
Pittsburgh, PA 15213

**John Lafferty**
School of Computer Science
Carnegie Mellon University
Pittsburgh, PA 15213

## Abstract

Recent research has made significant progress on the problem of bounding log partition functions for exponential family graphical models. Such bounds have associated dual parameters that are often used as heuristic estimates of the marginal probabilities required in inference and learning. However these variational estimates do not give rigorous bounds on marginal probabilities, nor do they give estimates for probabilities of more general events than simple marginals. In this paper we build on this recent work by deriving rigorous upper and lower bounds on event probabilities for graphical models. Our approach is based on the use of generalized Chernoff bounds to express bounds on event probabilities in terms of convex optimization problems; these optimization problems, in turn, require estimates of generalized log partition functions. Simulations indicate that this technique can result in useful, rigorous bounds to complement the heuristic variational estimates, with comparable computational cost.

## 1 Introduction

Undirected graphical models are natural and widely used in many domains, from statistical physics and image processing to social networks and contingency table analysis. For such models the log partition function, which arises in the exponential family representation, plays a fundamental role in most aspects of inference and learning. Although the log partition function is in general intractable to compute exactly, recent research has made considerable progress in obtaining effective bounds, opening up new possibilities for the further development and application of this important class of graphical models.

Variational methods and convex optimization have been the primary tools used in obtaining bounds on the log partition function. A key aspect of this method is that the bounds often have associated dual parameters, and these parameters can be used as heuristic estimates of the marginal probabilities required in inference and learning. Unfortunately, there is currently a gap in our understanding of how such dual parameters can be quantitatively related to the parameters of actual interest in the graphical model. In particular, the variational estimates do not give rigorous bounds on marginal probabilities, nor do they give estimates for probabilities of more general events than simple marginals.

In this paper we build upon this recent work by deriving rigorous upper and lower bounds on event probabilities for undirected graphical models. Our approach is based on the use of generalized Chernoff bounds to express bounds on event probabilities in terms of convex optimization problems. In the classical Chernoff bounding technique for independent and identically distributed random variables, linear bounds via the Markov inequality are further approximated in order to obtain analytic expressions for tail inequalities. As observed by Boyd and Vandenberghe (2004), this basic technique can be considerably generalized to obtain general event probability bounds using convex optimization. Although the idea behind such bounds is thought to be widely known, it has not been widely exploited (Stephen Boyd, personal communication).

Like their classical predecessors, generalized Chernoff bounds are obtained by minimizing a parameterized family of upper bounds—thus, they are naturally thought of as variational bounds. However, in our application of the Chernoff bound idea to graphical models, we introduce additional variational bounds to approximate log partition functions. We thus refer to the resulting bounds, somewhat redundantly for emphasis, as variational Chernoff bounds. For upper bounds we employ the approach of Wainwright and Jordan (2003b) based on semidefinite relaxations. In addition, we investigate the use of the tree-reweighted belief propagation algorithms of Wainwright et al. (2003), as well as a more direct approach based on barrier functions. For the lower bounds, available methods include structured mean field approximations and $M$-best approximations using belief propagation (Yanover & Weiss, 2003).

Classical Chernoff bounds are used almost exclusively for tail estimates of small probability events, where the courseness of the approximation does not affect asymptotic analysis. By using numerical optimization, the generalized approach has the potential to obtain much tighter bounds, which may be useful for more general events, and for smaller sample sizes. We carry out experiments with small graphical models which indicate that this technique can indeed result in effective bounds. Moreover, the bounds are obtained at comparable computational cost to the heuristic variational estimates. The technique thus provides a new tool for graphical modeling, giving rigorous bounds to complement the more heuristic estimates of mean parameters for which variational methods have been previously employed.

The remainder of the paper is organized as follows. In the following section we establish some notation and review some basic definitions associated with convex analysis and exponential representations of graphical models. In Section 3 we explain the idea behind generalized Chernoff bounds, and express the corresponding bounds for graphical models in terms of the log partition function. In this section we also review the classical Chernoff bounds, and give an example of how the generalized bounds apply to Markov and hidden Markov models as a simple special case. In Section 4 the variational approximations to log partition functions are used to derive manageable optimization problems. In Section 5 we show that the general bounds are, in fact, exact in certain cases. In Section 6 the results of simulations are presented that indicate the bounds can give nontrivial estimates of marginal probabilities, showing that the framework is not only appropriate for tail probabilities.

## 2 Background and Basic Definitions

We begin by establishing some notation and reviewing the basic properties of undirected graphical models and exponential family representations. We then review some basic concepts from convex analysis that are required.

### 2.1 Exponential family representations

Let $X = (X_1, X_2, \ldots, X_n)$ denote a random variable with $n$ components indexed by the nodes in a graph $G = (V, E)$, where each $X_i$ takes values in a finite set $\mathcal{X}$. We assume that $X$ has an exponential family distribution of the form

$$p_\theta(x) = \exp\left(\sum_\alpha \langle \theta, \phi(x) \rangle - \Phi(\theta)\right) \quad (1)$$

where $\langle \theta, \phi(x) \rangle = \sum_\alpha \theta_\alpha \phi_\alpha(x)$, and $\phi(x)$ is the vector of sufficient statistics. The index set $I = \{\alpha\}$ is determined by the clique structure of the graph; for example, we may have $\alpha = (s, t; i, j)$ corresponding to an indicator function $\phi_\alpha = \delta(x_s, i)\,\delta(x_t, j)$ for an edge $(s, t) \in E$. In a nonminimal representation, there are dependencies among the functions $\phi_\alpha$, and the number of parameters $m = |I|$ is larger than the dimension of the model.

The log partition function $\Phi(\theta)$ is the logarithm of the normalizing constant of the model, and is a convex function of $\theta$ satisfying $\partial \Phi(\theta)/\partial \theta_\alpha = E_\theta[\phi_\alpha(X)]$. The convex conjugate $\Phi^*$ is defined by $\Phi^*(\mu) = \sup_{\theta \in \mathbb{R}^m} \langle \mu, \theta \rangle - \Phi(\theta)$. If $\widehat{\theta} = \theta(\mu)$ is the parameter attaining the supremum, a calculation shows that $\Phi^*(\mu)$ can be expressed as a negative entropy $\Phi^*(\mu) = \sum_x p(x \,|\, \widehat{\theta}) \log p(x \,|\, \widehat{\theta})$ and $\mu_\alpha = E_{\widehat{\theta}}[\phi_\alpha(X)]$. These relations show that the dual parameters $\mu$ are the set of vectors that can be realized as marginals of $\phi$. The collection of such dual parameters is the *marginal polytope*

$$\mathrm{MARG}(G, \phi) = \quad (2)$$
$$\left\{\mu \in \mathbb{R}^m \,\middle|\, \sum_x p(x \,|\, \theta)\,\phi(x) = \mu \text{ for some } \theta \in \mathbb{R}^m\right\}$$

and plays a central role in the analysis of $\Phi(\theta)$. Since $\mathcal{X}$ is finite, the closure of $\mathrm{MARG}(G, \phi)$ is a finite intersection of halfspaces, and is thus indeed a polytope. It can be shown that

$$\Phi(\theta) = \sup_{\mu \in \mathrm{MARG}(G,\phi)} \langle \theta, \mu \rangle - \Phi^*(\mu) \quad (3)$$
$$= \sup_{\mu \in \mathcal{M}(\phi)} \langle \theta, \mu \rangle - \Phi^*(\mu) \quad (4)$$

where $\mathcal{M}(\phi) = \{\mu \in \mathbb{R}^n \,|\, \sum_x \phi(x)p(x) = \mu$ for some $p\}$. We refer to (Wainwright & Jordan, 2003a) for a comprehensive introduction to these constructions and their relevance to variational approximations.

### 2.2 Conjugacy and support functions

We now recall some basic definitions and conventions from convex analysis, referring to (Rockafellar, 1970) for further detail. Since generalized Chernoff bounds are based on linear approximations of convex (or possibly non-convex) sets, the notion of support function arises naturally. The *indicator function* $\delta_C$ of a set is defined as

$$\delta_C(x) = \begin{cases} 0 & \text{if } x \in C \\ \infty & \text{otherwise} \end{cases} \quad (5)$$

The *support function* of $C \subset \mathbb{R}^m$ is defined as

$$\delta_C^*(\lambda) = \sup_{x \in C} \langle x, \lambda \rangle \quad (6)$$

The suggestive notation comes from the fact that if $C$ is convex then the support function $\delta^*$ is in fact the convex conjugate of the the indicator function. If $C$ is convex, then $x$ lies in the closure $\mathrm{cl}\,C$ if and only if $\langle x, \lambda \rangle \leq \delta_C^*(\lambda)$ for every $\lambda$. In this case $(\delta_C^*)^* = \delta_{\mathrm{cl}\,C}$. This shows that a closed convex set $C$ can be represented as the solution set of a family of linear inequalities, and thus the support function characterizes $C$. We will also denote the support function by $S_C(\lambda)$.

For an undirected graphical model with parameter $\theta$, vector of sufficient statistics $\phi(x)$ and log partition function $\Phi(\theta)$, we will denote by $\Phi(f, \theta)$ the log partition function for the (generally non-graphical) model with probabilities proportional to $\exp(\langle \theta, \phi(x) \rangle + f(x))$; thus,

$$\Phi(f, \theta) = \sum_x \exp(\langle \theta, \phi(x) \rangle + f(x)) \qquad (7)$$

As a special case that will be useful below,

$$\log p(x \in C \mid \theta) = \Phi(-\delta_C, \theta) - \Phi(\theta) \qquad (8)$$

We will also use the notation $\Phi(-\delta_C, \theta) = \Phi_C(\theta)$.

## 3 Classical and Generalized Chernoff Bounds

Let $X$ be a real-valued random variable with distribution determined by some parameter $\theta$. The Markov inequality implies that for any $\lambda > 0$,

$$p_\theta(X \geq u) = p_\theta\left(e^{\lambda X} \geq e^{\lambda u}\right) \qquad (9)$$
$$\leq E_\theta[e^{\lambda(X-u)}] \qquad (10)$$

From this it follows that

$$\log p_\theta(X \geq u) \leq \inf_{\lambda \geq 0} \left(-\lambda u + \log E_\theta\left[e^{\lambda X}\right]\right) \qquad (11)$$

In the classical formulations of Chernoff bounds that are so widely used in probabilistic analysis, the relation (11) is further manipulated so that the upper bound has an analytic form. For example, if the random variable is $X \sim \text{Binomial}(n, p)$, it can easily be shown (see below) that

$$p_\theta\left(X < np(1-\delta)\right) \leq e^{-n\delta^2/2} \qquad (12)$$

Boyd and Vandenberghe (2004) observe that the basic idea behind inequality (11) can be considerably generalized in a way that involves convex optimization. Rather than deriving an expression that can be used to reason about tail probabilities analytically, one expresses an upper bound on the desired probability in terms of a convex optimization problem, and obtains a rigorous numerical bound on the probability by solving the optimization problem.

Let $X$ now denote a $\mathbb{R}^m$-valued random variable, whose distribution is indicated by a parameter $\theta$, and let $C \subset \mathbb{R}^m$. To bound the probability $p_\theta(X \in C)$, consider a parameterized family of upper bounds on the indicator function $-\delta_C$; that is, let $f_\lambda(x) \geq 0$ if $x \in C$. Then clearly

$$p_\theta(X \in C) \leq \inf_\lambda E_\theta\left[e^{f_\lambda}\right] \qquad (13)$$

In case $f_\lambda = \langle \lambda, x \rangle + u$ is affine, where $\lambda$ and $u$ are chosen subject to the constraint that $\langle \lambda, x \rangle + u \geq 0$ for $x \in C$, we have that

$$\log p_\theta(X \in C) \leq \inf_{\lambda, u} \log E_\theta\left[e^{\langle \lambda, x \rangle + u}\right] \qquad (14)$$
$$= \inf_{\lambda, u} \left(u + \log E_\theta\left[e^{\langle \lambda, x \rangle}\right]\right) \qquad (15)$$

Now, since $u \geq \langle -\lambda, x \rangle - \delta_C(x)$, it follows that $\inf u = \sup_x \langle -\lambda, x \rangle - \delta_C(x) = \delta_C^*(-\lambda)$. Therefore,

$$\log p_\theta(X \in C) \leq \inf_\lambda \left(\delta_C^*(-\lambda) + \log E_\theta\left[e^{\langle \lambda, x \rangle}\right]\right) \qquad (16)$$

For exponential family models, this line of argument leads to the following bounds.

**Proposition 1.** *Suppose that $X = (X_1, \ldots, X_m)$ is an exponential model with (non-minimal) sufficient statistic $\phi(x) \in \mathbb{R}^n$, and let $C \subset \mathbb{R}^m$. Then*

$$\log p_\theta(X \in C) = \Phi(-\delta_C, \theta) - \Phi(\theta) \qquad (17)$$
$$\leq \inf_\lambda \Phi(f_\lambda, \theta) - \Phi(\theta) \qquad (18)$$

*for any family of functions $f_\lambda \geq -\delta_C$ bounding the indicator function. In particular,*

$$\log p_\theta(X \in C) \leq \inf_{\lambda \in \mathbb{R}^n} S_{C,\phi}(-\lambda) + \Phi(\lambda + \theta) - \Phi(\theta) \qquad (19)$$

*where $S_{C,\phi}(y) = \sup_{x \in C} \langle y, \phi(x) \rangle$, for $y \in \mathbb{R}^n$.*

*Proof.* The equality in (17) follows from

$$\log p_\theta(X \in C) = \log \sum_{x \in C} e^{\langle \theta, \phi(x) \rangle} - \Phi(\theta)$$
$$= \log \sum_x e^{-\delta_C(x) + \langle \theta, \phi(x) \rangle} - \Phi(\theta)$$
$$= \Phi_C(\theta) - \Phi(\theta)$$

Let $f_\lambda(x) = \langle \lambda, \phi(x) \rangle + u$ be an affine upper bound on $-\delta_C$. Then following the argument above, the bound in (19) follows from observing that $\log E_\theta\left[e^{\langle \lambda, \phi(X) \rangle}\right] = \Phi(\lambda + \theta) - \Phi(\theta)$. □

In case the vector of sufficient statistics includes each $X_i$, by restricting the linear function to one of the form $f_\lambda = \langle \lambda, x \rangle + u$ rather than $f_\lambda = \langle \lambda, \phi(x) \rangle + u$, we obtain a generally weaker bound of the form

$$\log p_\theta(X \in C) \leq \inf_{\lambda \in \mathbb{R}^m} S_C(-\lambda) + \Phi(\lambda + \theta) - \Phi(\theta) \qquad (20)$$

where now $S_C = \delta_C^*$ is the standard support function.

### 3.1 Classical Chernoff bounds

Classical Chernoff bounds (Chernoff, 1952; Motwani & Raghavan, 1995) are widely used to obtain rough, analytically convenient bounds on tail probabilities for iid observations. If $X_1, X_2, \ldots, X_n$ are independent Bernoulli$(p)$

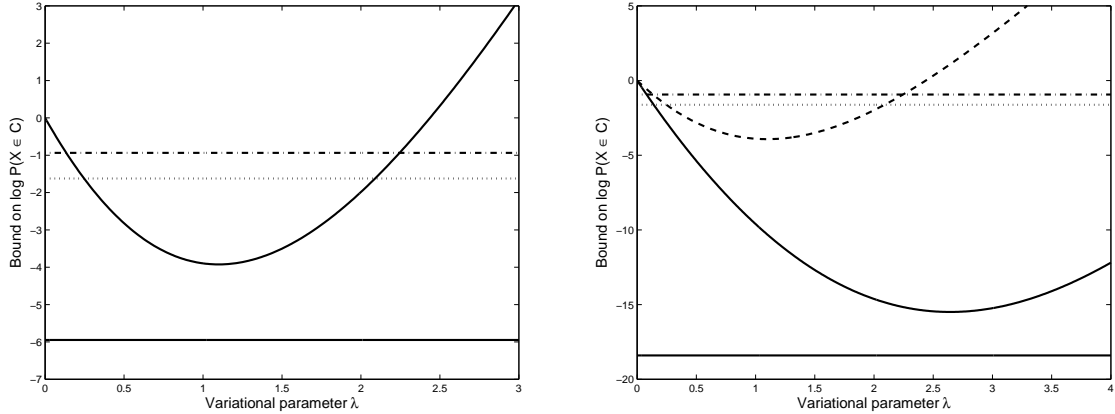

Figure 1: Classical and optimized Chernoff bounds for independent Bernoulli trials (left) and a Markov model (right) for $C_\delta = \{X \mid \sum X_i > np(1+\delta)\}$ with $p = \frac{1}{2}$ and $\delta = \frac{1}{2}$. Left: $n = 30$ Bernoulli trials—the classical Chernoff bound $\log P(X \in C_\delta) < -np\delta^2/4$ (top horizontal line), $\log P(X \in C_\delta) < np(\delta - (1+\delta)\log(1+\delta))$ (second horizontal line), and true probability (lower horizontal line); the curve shows the variational approximation $\log P(X \in C_\delta) < -\lambda np(1+\delta) - \Phi(\theta + \lambda) - \Phi(\theta)$. Right: bounds for a Markov model with $n = 30$, $\theta_{1,1} = -1$ and $\theta_1 = \log p/(1-p)$. The curved line is the variational approximation, where the log partition functions are computed exactly using dynamic programming; the bottom horizontal line is the true probability. The dashed curve is the variational approximation that assumes independent $X_i$ (same as curve in left plot).

trials, the upper Chernoff bound is established by using the Markov inequality to obtain

$$\log p(X \in C_\delta) \leq \inf_\lambda \left( -\lambda np(1+\delta) + E\left[e^{\lambda \sum_i X}\right] \right)$$

for $C_\delta = \{X \mid \sum_i X_i \geq np(1+\delta)\}$; this is equivalent to using the linear approximation to the indicator function employed above. Using the convexity of $\exp$, in the form $1 - x < e^{-x}$, the moment generating function $E[e^{\lambda \sum_i X_i}]$ is bounded from above as

$$\log E[e^{\lambda \sum_i X_i}] = \sum_i \log\left(e^\lambda p + 1 - p\right) \quad (21)$$
$$\leq np(1 - e^\lambda) \quad (22)$$

This upper bound is then minimized to obtain the optimal $\lambda = \log(1+\delta)$, and thus the Chernoff bound

$$p(X \in C_\delta) \leq \left(\frac{e^\delta}{(1+\delta)^{(1+\delta)}}\right)^{np} \quad (23)$$

A more commonly used form, because of its simplicity, is the weaker bound

$$p(X \in C_\delta) \leq e^{-np\delta^2/4} \quad (24)$$

which is valid when $\delta < 2e - 1$.

We review this elementary analysis to point out that our approach in the general case for exponential family graphical models is parallel. A family of upper bounds is expressed using a linear approximation to the indicator function, and this family of upper bounds is expressed in terms of the log partition function (or moment generating function). The log partition function is then approximated by an upper bound that is more amenable to computation—however, in the general case, such computation will involve more sophisticated approximations and convex optimization.

### 3.2 Example: Chernoff bounds for Markov models

One of the simplest extensions of the classical Chernoff bounds for independent Bernoulli trials is the case of a Markov or hidden Markov model. For illustration we consider a Markov model on two states, where the joint distribution for $X_1, \ldots, X_m$ with $X_i \in \{0, 1\}$ is given by

$$p(X_1, \ldots, X_n) \propto \exp\left(\sum_{i=1}^n \theta_{X_i} + \sum_{i=1}^{m-1} \theta_{X_i, X_{i+1}}\right) \quad (25)$$

Thus $\theta = (\theta_0, \theta_1, \theta_{0,0}, \theta_{0,1}, \theta_{1,0}, \theta_{1,1})$, with the case $\theta_{0,0} = \theta_{0,1} = \theta_{1,0} = \theta_{1,1} = 0$ corresponding to independent Bernoulli($p$) trials with $p = e^{\theta_1}/(e^{\theta_0} + e^{\theta_1})$.

Since the random variables are not independent, the classical Chernoff bound for $p_\theta(\sum_i X_i > np(1+\delta))$ will be highly biased. The generalized Chernoff bound for the event $C_\delta = \{X \mid \sum_i X_i \geq np(1+\delta)\}$ is

$$p_\theta(X \in C_\delta) \leq \inf_\lambda -\lambda np(1+\delta) + \Phi(\theta + \bar\lambda) - \Phi(\theta)$$

where $\bar\lambda = (0, \lambda, 0, 0, 0, 0)$. In this case the log partition functions $\Phi(\theta + \lambda)$ and $\Phi(\theta)$ are easily computed in $O(n)$ time using dynamic programming. However, computing the probability $p_\theta(X \in C_\delta)$ exactly using dynamic programing requires $O(n^2)$ time—auxiliary states to count $\sum_i X_i$ must be introduced, requiring $O(n)$ states at each position. Similar statements can be made for graphical models where the underlying graph is a tree.

An example of these bounds for a simple Markov model is shown in Figure 1, where the bounds are compared to the classical bounds for the iid case. The left plot shows bounds for Bernoulli trials with $p = \frac{1}{2}$; the right plot shows bounds for a Markov model of the form (25) with $\theta_1 = \log \frac{p}{1-p}$ and $\theta_{1,1} = -1$, which discourages neighboring 1s.

Such a tree-based graphical model is the simplest case of the generalized Chernoff bounds we consider. For more general graphical models, where dynamic programming may not be available, we must resort to more elaborate approximations.

## 4 Variational Chernoff Bounds

The exact log probability (17) and the generalized Chernoff bounds (19) require computation of log partition functions. In order to derive tractable bounds, we apply upper and lower variational bounds. Let $\Phi^{(U)}(\theta)$ and $\Phi^{(L)}(\theta)$ be upper and lower bounds on $\Phi(\theta)$, respectively. Then clearly

$$\log p_\theta(X \in C) \leq \Phi_C^{(U)}(\theta) - \Phi^{(L)}(\theta) \quad (26)$$
$$\log p_\theta(X \in C) \geq \Phi_C^{(L)}(\theta) - \Phi^{(U)}(\theta) \quad (27)$$

and, in addition, applying the bounds to (19) gives

$$\log p_\theta(X \in C) \leq \quad (28)$$
$$\inf_{\lambda \in \mathbb{R}^n} S_{C,\phi}(-\lambda) + \Phi^{(U)}(\lambda + \theta) - \Phi^{(L)}(\theta)$$

In this section we describe the application of semidefinite relaxations and tree-based belief propagation in this framework.

### 4.1 Semidefinite relaxation

Wainwright and Jordan (2003b) develop a semidefinite relaxation of $\Phi(\theta)$ which leads to a log determinant optimization problem. The idea behind this approach is to bound the dual function $\Phi^*$, which is a negative entropy, in terms of the entropy of a Gaussian. Since the entropy of a Gaussian is a log determinant, the semidefinite upper bound follows. The analysis in (Wainwright & Jordan, 2003b) is restricted to the case of $\mathcal{X} = \{-1, 1\}$ and vertex and pairwise interaction potentials on the complete graph $K_n$; this is the case we now assume, although the approach generalizes.

Recalling some of the notation of (Wainwright & Jordan, 2003b), for $\mu \in \mathbb{R}^m$, $M_1[\mu]$ is the $(n+1) \times (n+1)$ matrix

$$M_1[\mu] = \begin{pmatrix} 1 & \mu_1 & \cdots & \mu_n \\ \mu_1 & 1 & \cdots & \mu_{1n} \\ \mu_2 & \mu_{21} & \cdots & \mu_{2n} \\ \vdots & \vdots & \vdots & \vdots \\ \mu_{n-1} & \mu_{(n-1),1} & \cdots & \mu_{(n-1),n} \\ \mu_n & \mu_{n1} & \cdots & 1 \end{pmatrix} \quad (29)$$

and $\text{SDEF}_1(K_n) = \{\mu \mid M_1[\mu] \succeq 0\}$.

**Proposition 2.** *Let* $M \supset \text{MARG}(K_n)$ *be any convex set that contains* $\text{SDEF}_1(K_n)$. *Then*

$$\log p_\theta(X \in C) \leq \quad (30)$$
$$\sup_{\mu \in M} \left\{ \langle \theta, \mu \rangle + \frac{1}{2} \log \det A(\mu) - S_{C,\phi}^*(\mu) \right\} + c_n - \Phi(\theta)$$

*with* $A(\mu) = M_1[\mu] + \frac{1}{3}\widetilde{I}$, *where* $\widetilde{I} = [0, I_n]$ *is an* $(n+1) \times (n+1)$ *block diagonal matrix, and* $c_n = \frac{n}{2} \log \left(\frac{\pi e}{2}\right)$.

The proof of this proposition follows from inequality (28) and Theorem 1 of (Wainwright & Jordan, 2003b) after observing that $S_{C,\phi}(\lambda)$, as a supremum of linear functions, is a convex function even if $C$ is not convex, and that

$$S_{C,\phi}^*(\mu) = \sup_\lambda \langle \lambda, \mu \rangle - S_{C,\phi}(\lambda) \quad (31)$$
$$= -\inf_\lambda \langle \lambda, \mu \rangle + S_{C,\phi}(-\lambda) \quad (32)$$

In particular, $-S_{C,\phi}^*(\mu)$ is a concave function of $\mu$. Thus, solving the log determinant optimization problem above and replacing $\Phi(\theta)$ with any lower bound $\Phi^{(L)}(\theta)$ gives an upper bound on $p_\theta(X \in C)$.

### 4.2 Tree-reweighted belief propagation

An alternative approach to obtaining upper bounds on $\Phi(\theta + \lambda)$ is based on belief propagation. Wainwright et al. (2003) show that $\Phi(\theta + \lambda)$ can be bounded from above by minimizing a certain functional over "pseudomarginals" on the nodes and edges in the graph that satisfy certain consistency constraints. In order to use this method to obtain upper bounds on $\log p_\theta(X \in C)$, it is necessary to minimize over $\lambda$. This is possible using a kind of pseudo-gradient descent algorithm that mirrors the approximate maximum likelihood estimation given in (Wainwright et al., 2003); we do not pursue this approach here. For simplicity, in the experiments reported in Section 6 we instead use the tree-reweighted BP algorithm to derive *lower* bounds on $\log p_\theta(X \in C)$ using inequality (27).

However, the tree-based approach is somewhat limited. Since $\Phi(-\delta_C, \theta) = \log \sum_x \exp(\langle \theta, \phi(x) \rangle - \delta_C(x))$, we see that $\delta_C(x)$ may introduce additional couplings among the nodes, forcing one to use more complicated variational methods. For example, if $\phi(x)$ contains only node and edge potentials, then we can use normal tree-based variational methods to compute a bound for $\Phi(\theta)$. However, if $\delta_C(x)$ introduces coupling of more than two nodes, then one cannot use simple tree-based methods anymore, but would have to resort to hyper-tree based methods to compute a bound for $\Phi(-\delta_C, \theta)$.

### 4.3 Lower bounds on $\Phi$

At least two techniques are available for the required lower bounds $\Phi^{(L)}(\theta)$. By conjugate duality we have that

$$\Phi(\theta) = \sup_{\mu \in \text{MARG}(G,\phi)} (\langle \theta, \mu \rangle - \Phi^*(\mu)) \quad (33)$$

$$\geq \max_{\mathcal{M}_{\text{tract}}(G,\phi)} (\langle \theta, \mu \rangle - \Phi^*(\mu)) \quad (34)$$

where $\mathcal{M}_{\text{tract}}(G,\phi) \subset \text{MARG}(G,\phi)$ is any subset contained within the marginal polytope. The structured mean field approximation adopts a tractable subset of smaller dimension than $\text{MARG}(G,\phi)$, for which the max can be carried out using efficient iterative algorithms. However, $\mathcal{M}_{\text{tract}}(G,\phi)$ is typically not convex, and the iterative algorithms generally suffer from the presence of many local maxima; see (Wainwright & Jordan, 2003a) for an overview.

An alternative approach is to make the approximation

$$\Phi(\theta) \geq \log \left( \sum_{x \in M\text{-Best}} \exp \left( \langle \theta, \phi(x) \rangle \right) \right) \quad (35)$$

where $M$-best is a set of (approximately) most probable configurations. Yanover and Weiss (2003) develop an algorithm based on loopy belief propagation to efficiently compute an approximate $M$-best set. Such an approximation is expected to be good when there are a few highly probable configurations. We have obtained good results with this approach, but report below on the use of the structured mean field approximation.

## 5 Tightness of Chernoff Bounds

In this section we show that the generalized Chernoff bounds with linear approximations to the indicator function are actually *exact* expressions of event probabilities in an exponential family graphical model in certain cases. While the actual computation of the Chernoff bounds may be highly nontrivial, this result gives an indication of the power of the framework.

**Proposition 3.** *Let $p_\theta(X) = \exp(\langle \theta, \phi(X) - \Phi(\theta) \rangle)$ be an exponential model with $X = (X_1, \ldots, X_m)$, where $X \mapsto \phi(X) \in \mathbb{R}^n$ is a one-to-one mapping. Then for $C \subset \mathbb{R}^m$,*

$$\log p_\theta(X \in C) = \inf_{\lambda \in \mathbb{R}^n} S_{C,\phi}(-\lambda) + \Phi(\lambda + \theta) - \Phi(\theta)$$

Thus the inequality in (19) is in fact an equality. In order to show this we first give two lemmas. Recall that $\mathcal{M}(\phi)$ is the polytope of mean parameters associated with $\phi$. Define $\mathcal{M}_C(\phi)$ to be the mean parameters over probabilities restricted to $C$: $\mathcal{M}_C(\phi) = \{\mu \in \mathbb{R}^n \mid \sum_{x \in C} p(x) \phi(x) = \mu$ for some $p$ with $\sum_{x \in C} p(x) = 1\}$. The first lemma we state without proof, referring to (Rockafellar, 1970) for details on support functions.

**Lemma 1.** $S^*_{C,\phi}(\mu)$ is the indicator function $\delta_{\mathcal{M}_C(\phi)}$.

**Lemma 2.** *For $\mu \in \mathcal{M}_C(\phi)$, $\Phi^*(\mu) = \Phi^*_C(\mu)$.*

*Proof.* Let $\mu \in \mathcal{M}_C$, with $\mu = \sum_x \phi(x) q(x)$ and $\sum_{x \in C} q(x) = 1$. Suppose that $\mu \in \text{bd}\mathcal{M}$. Then since $\mathcal{M}$ is closed and $\theta \mapsto \Lambda(\theta) = \sum_x \phi(x) p(x \mid \theta)$ is onto $\text{ri}\mathcal{M}$ (Wainwright & Jordan, 2003a), there exists a sequence $\mu_n \in \text{ri}\mathcal{M}$ with $\mu_n = \sum_x \phi(x) p(x \mid \theta_n)$, where $p(x \mid \theta_n) \to q(x)$ and $\mu_n \to \mu$. Thus, $\lim_n p(x \mid \theta_n) = 0$ if $x \notin C$, and hence $\lim \Phi(\theta_n) = \lim \Phi_C(\theta_n)$. Since $\Phi^*(\mu_n) = \sup_\theta (\langle \theta, \mu_n \rangle - \Phi(\theta)) = \langle \theta_n, \mu_n \rangle - \Phi(\theta_n)$, for this optimal value of $\theta_n$ we have that

$$\lim_n \Phi^*(\mu_n) = \lim_n \langle \theta_n, \mu_n \rangle - \Phi(\theta_n)$$
$$= \lim_n \langle \theta_n, \mu_n \rangle - \Phi_C(\theta_n)$$
$$= \lim_n \Phi^*_C(\mu_n) = \Phi^*_C(\mu)$$

The analysis for $\mu \in \text{ri}M$ is similar. □

*Proof of Proposition 3.* Define

$$h(\mu, \lambda; \theta) = S_{C,\phi}(-\lambda) + \langle \theta + \lambda, \mu \rangle - \Phi^*(\mu) - \Phi(\theta)$$

Then the Chernoff bound (19) can be expressed as

$$\log p_\theta(X \in C) \leq \inf_\lambda S_{C,\phi}(-\lambda) + \Phi(\theta + \lambda) - \Phi(\theta)$$
$$= \inf_\lambda \sup_{\mu \in \mathcal{M}} S_C(-\lambda) + \langle \theta + \lambda, \mu \rangle - \Phi^*(\mu) - \Phi(\theta)$$
$$= \inf_\lambda \sup_{\mu \in \mathcal{M}} h(\mu, \lambda; \theta)$$

Now, reversing the sup and inf, we have that

$$\sup_{\mu \in \mathcal{M}} \inf_\lambda h(\mu, \lambda; \theta)$$
$$= \sup_{\mu \in \mathcal{M}} \inf_\lambda S_{C,\phi}(-\lambda) + \langle \theta + \lambda, \mu \rangle - \Phi^*(\mu) - \Phi(\theta)$$
$$= \sup_{\mu \in \mathcal{M}} \langle \theta, \mu \rangle - \Phi^*(\mu) - S^*_{C,\phi}(\mu) - \Phi(\theta)$$
$$= \sup_{\mu \in \mathcal{M}_C} \langle \theta, \mu \rangle - \Phi^*_C(\mu) - \Phi(\theta)$$
$$= \Phi_C(\theta) - \Phi(\theta) = \log p_\theta(X \in C)$$

where the third equality follows from Lemma 1 and the fourth equality follows from Lemma 2.

Now, note that since $S_{C,\phi}(-\lambda)$ is convex in $\lambda$, $\langle \theta + \lambda, \mu \rangle - \Phi^*(\mu)$ is concave in $\mu$, the marginal polytope $\mathcal{M}(\phi)$ is convex and compact, and $-\Phi^*(\mu)$ and consequently $\langle \theta + \lambda, \mu \rangle - \Phi^*(\mu)$ are upper semicontinuous on $\mathcal{M}$ (Wainwright & Jordan, 2003a), we can conclude that $\sup_{\mu \in \mathcal{M}} \inf_\lambda h(\mu, \lambda; \theta) = \inf_\lambda \sup_{\mu \in \mathcal{M}} h(\mu, \lambda; \theta)$ from standard minimax results (Peck & Dumage, 1957). □

| Problem type | | | Average $L_1$ error $\pm$ std | | | |
|---|---|---|---|---|---|---|
| | | | Approximation method | | | |
| Graph | Coupling | Strength | MF/Tree lower | MF/SDP lower | Tree/MF upper | SDP heuristic |
| Grid | Repulsive | (0.25,1.0) | $0.093 \pm 0.003$ | $0.297 \pm 0.009$ | $0.166 \pm 0.008$ | $0.010 \pm 0.002$ |
| | Repulsive | (0.25,2.0) | $0.127 \pm 0.009$ | $0.290 \pm 0.007$ | $0.327 \pm 0.059$ | $0.024 \pm 0.002$ |
| | Mixed | (0.25,1.0) | $0.054 \pm 0.028$ | $0.452 \pm 0.047$ | $0.070 \pm 0.038$ | $0.026 \pm 0.002$ |
| | Mixed | (0.25,2.0) | $0.095 \pm 0.012$ | $0.421 \pm 0.053$ | $0.138 \pm 0.011$ | $0.017 \pm 0.003$ |
| | Attractive | (0.25,1.0) | $0.026 \pm 0.001$ | $0.770 \pm 0.019$ | $0.025 \pm 0.002$ | $0.023 \pm 0.001$ |
| | Attractive | (0.25,2.0) | $0.001 \pm 0.001$ | $0.791 \pm 0.026$ | $0.001 \pm 0.001$ | $0.016 \pm 0.002$ |
| Full | Repulsive | (0.25,0.25) | $0.072 \pm 0.010$ | $0.290 \pm 0.006$ | $0.069 \pm 0.011$ | $0.021 \pm 0.001$ |
| | Repulsive | (0.25,0.50) | $0.132 \pm 0.009$ | $0.238 \pm 0.007$ | $0.156 \pm 0.016$ | $0.016 \pm 0.001$ |
| | Mixed | (0.25,0.25) | $0.032 \pm 0.001$ | $0.393 \pm 0.014$ | $0.029 \pm 0.001$ | $0.013 \pm 0.004$ |
| | Mixed | (0.25,0.50) | $0.120 \pm 0.027$ | $0.450 \pm 0.037$ | $0.127 \pm 0.034$ | $0.024 \pm 0.004$ |
| | Attractive | (0.25,0.06) | $0.009 \pm 0.001$ | $0.445 \pm 0.009$ | $0.007 \pm 0.001$ | $0.019 \pm 0.003$ |
| | Attractive | (0.25,0.12) | $0.037 \pm 0.006$ | $0.520 \pm 0.023$ | $0.033 \pm 0.006$ | $0.040 \pm 0.003$ |

Table 1: $L_1$ approximation error of single node marginals for the fully connected graph $K_9$ and the 4 nearest neighbour grid with 9 nodes, with varying potential and coupling strengths $(d_{\text{pot}}, d_{\text{coup}})$. Three different variational methods are compared: MF/Tree derives a lower bound with mean field approximation for $\Phi_C$ and tree-reweighted belief propagation for $\Phi$; MF/SDP derives a lower bound with the SDP relaxation used for $\Phi$; Tree/MF derives an upper bound using tree-reweighted belief propagation for $\Phi_C$ and mean field for $\Phi$. SDP denotes the heuristic use of the dual parameters in the SDP relaxation, with no provable upper or lower bounds.

## 6 Experimental Results

To test the performance of the upper and lower bound methods, we performed experiments for binary random fields on both a complete graph and a 2-D nearest-neighbor grid graph, closely following the experiments in (Wainwright & Jordan, 2003b). In order to be able to compare the bounds with the exact probabilities, we show results for small graphs with 9 nodes. For different qualitative characteristics of the exponential distributions (repulsive, mixed, or attractive), we construct many randomly generated models, and compute the mean error for each type of graph.

The graphical models were randomly generated according to the following specification. First, the parameters were randomly generated in the following manner:

*Single node potentials*: For each trial, we sample $\theta_s \sim$ Uniform$(-d_{\text{pot}}, +d_{\text{pot}})$ independently for each node, where $d_{\text{pot}} = \frac{1}{4}$.

*Edge coupling potentials*: For a given coupling strength $d_{\text{coup}}$, three types of coupling are used:

Repulsive: $\theta_{st} \sim$ Uniform$(-2d_{coup}, 0)$
Mixed: $\theta_{st} \sim$ Uniform$(-d_{\text{coup}}, +d_{\text{coup}})$
Attractive: $\theta_{st} \sim$ Uniform$(0, 2d_{\text{coup}})$

For a given model, the marginal probabilities $p_\theta(X_s = 1)$ and $p_\theta(X_s = 1, X_t = 1)$ are computed exactly for each node and edge by calculating the log partition function exactly. Then, the variational Chernoff bounds on these probabilities are computed using different approximations to the log partition functions. As described in Sections 3 and 4, we have that $\log p_\theta(X \in C) = \Phi_C(\theta) - \Phi(\theta)$. In the case of the marginal at a single node, $C = \{x \in \mathbb{R}^n \mid x_s = 1\}$. We compute the bounds using the following methods:

*MF/Tree*: A lower bound on $\log p_\theta(X \in C)$ is computed by applying the structured mean field approximation to $\Phi_C(\theta)$ and the tree-reweighted belief propagation approximation to $\Phi(\theta)$.

*MF/SDP*: A lower bound on $\log p_\theta(X \in C)$ is computed by the applying structured mean field approximation to $\Phi_C(\theta)$ and the semidefinite relaxation, resulting in a log determinant problem for $\Phi(\theta)$.

*Tree/MF*: An upper bound is derived using tree-reweighted belief propagation to upper bound $\Phi_C(\theta)$, and using structured mean field to derive a lower bound on $\Phi(\theta)$.

*SDP*: The semidefinite relaxation is used to heuristically estimate the marginal probability, as in (Wainwright & Jordan, 2003b), with no provable upper or lower bound.

To assess the accuracy of each approximation, we use the $L_1$ error, defined as

$$\frac{1}{n} \sum_{s=1}^{n} |p_\theta(X \in C) - \widehat{p}_\theta(X \in C)| \qquad (36)$$

where $\widehat{p}_\theta$ denotes the estimated marginal. The results are shown in Table 1 for the single node case, and in Table 2 for the case of node pairs.

| Problem type | | | Average $L_1$ error $\pm$ std | | | |
|---|---|---|---|---|---|---|
| | | | Approximation method | | | |
| Graph | Coupling | Strength | MF/Tree lower | MF/SDP lower | Tree/MF upper | SDP heuristic |
| Grid | Repulsive | (0.25,1.0) | $0.025 \pm 0.003$ | $0.118 \pm 0.012$ | $0.047 \pm 0.008$ | $0.005 \pm 0.003$ |
| | Repulsive | (0.25,2.0) | $0.034 \pm 0.005$ | $0.108 \pm 0.010$ | $0.101 \pm 0.022$ | $0.013 \pm 0.001$ |
| | Mixed | (0.25,1.0) | $0.026 \pm 0.004$ | $0.243 \pm 0.022$ | $0.037 \pm 0.009$ | $0.019 \pm 0.005$ |
| | Mixed | (0.25,2.0) | $0.056 \pm 0.024$ | $0.250 \pm 0.035$ | $0.087 \pm 0.031$ | $0.021 \pm 0.006$ |
| | Attractive | (0.25,1.0) | $0.029 \pm 0.008$ | $0.621 \pm 0.076$ | $0.043 \pm 0.015$ | $0.016 \pm 0.012$ |
| | Attractive | (0.25,2.0) | $0.002 \pm 0.001$ | $0.791 \pm 0.012$ | $0.003 \pm 0.001$ | $0.036 \pm 0.007$ |
| Full | Repulsive | (0.25,0.25) | $0.011 \pm 0.002$ | $0.081 \pm 0.024$ | $0.015 \pm 0.001$ | $0.021 \pm 0.004$ |
| | Repulsive | (0.25,0.50) | $0.008 \pm 0.005$ | $0.046 \pm 0.003$ | $0.021 \pm 0.002$ | $0.021 \pm 0.003$ |
| | Mixed | (0.25,0.25) | $0.040 \pm 0.006$ | $0.216 \pm 0.013$ | $0.014 \pm 0.001$ | $0.012 \pm 0.007$ |
| | Mixed | (0.25,0.50) | $0.068 \pm 0.011$ | $0.250 \pm 0.033$ | $0.052 \pm 0.005$ | $0.016 \pm 0.011$ |
| | Attractive | (0.25,0.06) | $0.020 \pm 0.004$ | $0.257 \pm 0.017$ | $0.003 \pm 0.001$ | $0.026 \pm 0.007$ |
| | Attractive | (0.25,0.12) | $0.061 \pm 0.009$ | $0.367 \pm 0.019$ | $0.015 \pm 0.003$ | $0.061 \pm 0.005$ |

Table 2: $L_1$ approximation error of pairwise node marginals. Approximation methods are as described for Table 1.

## 7 Conclusion

In this paper a framework for deriving rigorous bounds on probabilities for graphical models was proposed. Using generalized Chernoff bounds, the technique derives probability bounds in terms of convex optimization problems involving certain support functions and a difference of log partition functions. We showed that these bounds are in fact exact under certain conditions, which gives an indication of the power of the framework. Semidefinite relaxations and tree-reweighted belief propagation were used to derive tractable forms of the bounds. Experimental results on small graphs indicated that the approach can give useful bounds that are comparable, though generally weaker than, the heuristic estimates obtained using dual parameters, with tree-based approximations giving better accuracy than semidefinite relaxations.

Recent progress in bounding log partition functions has both enabled this work, and highlighted the need for rigorous bounds to complement the heuristic use of dual parameters in variational methods. Interesting directions for further developing this approach include the use of alternative approximations, such as spanning tree methods that permit estimates for more general events than simple marginals, and the application of the approach to tail probability estimates for complex models.